\pdfoutput=1

\documentclass[11pt]{article}

\usepackage[]{acl}

\usepackage{times}
\usepackage{latexsym}

\usepackage[T1]{fontenc}

\usepackage[utf8]{inputenc}

\usepackage{microtype}
\usepackage{paralist}
\usepackage{diagbox}
\usepackage{relsize}

\usepackage{booktabs}
\usepackage{amsmath}

\usepackage{cleveref}
\usepackage{graphicx}
\usepackage{multicol}
\usepackage{multirow}
\usepackage{array}
\usepackage{tcolorbox}
\usepackage{xspace}
\usepackage{pgf}
\usepackage{collcell}
\usepackage{subcaption}
\newcolumntype{P}[1]{>{\centering\arraybackslash}p{#1}}

\newcommand{\Oversegmented}[1]{\textcolor{red}{Over-segmented}}

\newcommand{\Undersegmented}[1]{\textcolor{yellow}{Under-segmented}}

\newcommand{\Correct}[1]{\textcolor{green}{Correct}}
\newcommand{\ex}[1]{\textit{#1}\xspace}
\newcommand{\x}{\phantom{0}}

\title{Revisiting subword tokenization: A case study on affixal negation in large language models}

\author{\textbf{Thinh Hung Truong$^{1}$ \quad Yulia Otmakhova$^{1}$ \quad Karin Verspoor$^{2,1}$} \\
    \textbf{Trevor Cohn$^{1,}$\thanks{\, Also at Google Research.} \quad Timothy Baldwin$^{3,1}$}\\[1ex]
$^1$The University of Melbourne\,\,\, $^2$RMIT University\,\,\,  $^3$MBZUAI\\[1ex]
\smaller \texttt{\{hungthinht,yotmakhova\}@student.unimelb.edu.au} \\ 
\smaller \texttt{karin.verspoor@rmit.edu.au}\qquad \texttt{trevor.cohn@unimelb.edu.au} \qquad \texttt{tb@ldwin.net}
}

\begin{document}

\maketitle

\begin{abstract}

In this work, we measure the impact of affixal negation on modern English large language models (LLMs). 
In affixal negation, the negated meaning is expressed through a negative morpheme, which is potentially challenging for LLMs as their tokenizers are often not morphologically plausible.
We conduct extensive experiments using LLMs with different subword tokenization methods, which lead to several insights on the interaction between tokenization performance and negation sensitivity. 
Despite some interesting mismatches between tokenization accuracy and negation detection performance, we show that models can, on the whole, reliably recognize the meaning of affixal negation.

\end{abstract}

\section{Introduction}

Negation is central to language understanding but is not properly captured by modern NLP methods \citep[inter alia]{hossain-etal-2022-analysis, truong-etal-2023-language}.
While state-of-the-art large language models (LLMs) have improved negation-related capabilities, challenges remain, such as the ability to correctly determine the enclosed scope of negation, or when negation interacts with other linguistic constructions like quantifiers \citep{she-etal-2023-scone, truong-etal-2023-language}.
Negations in common English NLP benchmarks are typically marked by separate negation cues such as \textit{not} or \textit{no}. 
However, in practice, negation can also be expressed through morphemes (or affixes) of words, i.e.\ by negative prefixes or suffixes such as in  \textit{uninteresting} or \textit{effortless}.  



\begin{figure}[!tbp]
    \centering
    \includegraphics[width = \columnwidth]{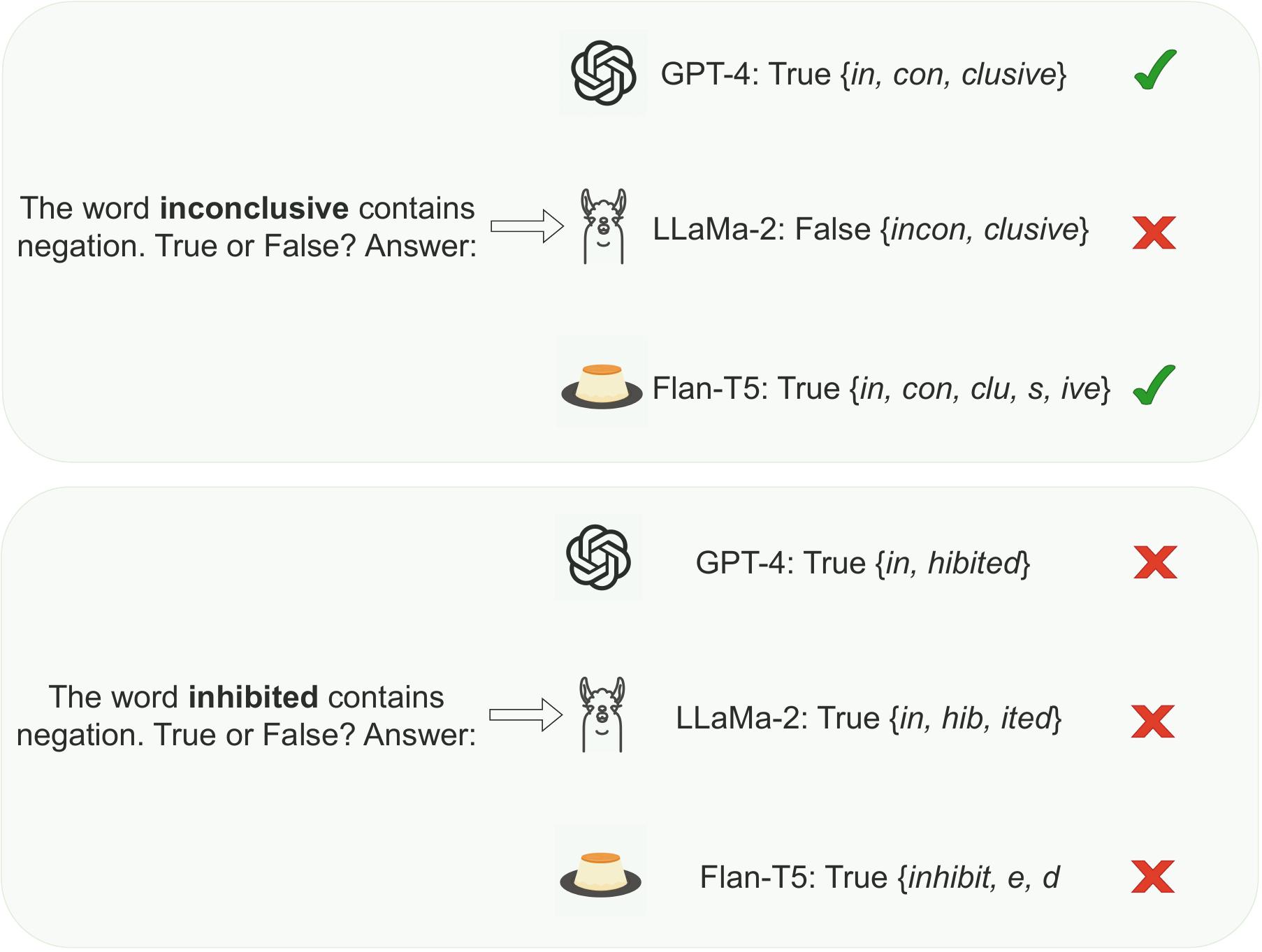}
    \caption{Example of our affixal negation prediction task, with the  tokenization output for each model.}
    \label{fig:example}
\end{figure}

\begin{table*}[!htbp]
\footnotesize
    \centering
    \begin{tabular}{lll p{3cm} cll}
    \toprule
      Model & Type & Variant & Data source & Vocab. size &  Output & NegMorph \\
        \midrule
        BERT & BPE & WordPiece & books, wiki &                                                 \x30K & \{\ex{anti, clin, al}\} & Correct \\
        RoBERTa & BPE & Byte-level BPE & books, wiki &                                         \x50K & \{\ex{antic, l, inal}\} & Under-segmented \\
        XLNet  & ULM & SentencePiece & book, wiki, web text &                                  \x32K & \{\ex{anti, clin, al}\} & Correct \\
        AlBERT & ULM & SentencePiece & book, wiki &                                            \x32K & \{\ex{anti, clin, al}\} & Correct \\
        T5 & BPE & SentencePiece & web text &                                                  \x32K & \{\ex{anti, clin, al}\} & Correct \\
        Llama-2 & BPE & SentencePiece & web text, code, books, wiki, scientific publications & \x32K & \{\ex{ant, ic, l, inal}\} & Over-segmented \\
        GPT-2 & BPE & Byte-level BPE & web text &                                              \x50K & \{\ex{antic, l, inal}\} & Under-segmented \\
        GPT-4 & BPE & Byte-level BPE & undisclosed &                                            100K & \{\ex{antic, l, inal}\} & Under-segmented \\
    \bottomrule
    \end{tabular}
    \caption{Summary of different tokenizers used in our experiments. Output are tokenized version of the word \ex{anticlinal} (model-specific special tokenization characters are removed for clarity purposes). All models are the base version unless otherwise specified.}
    \label{tab:summary}
\end{table*}

While humans can identify affixal negation by leveraging morphological cues, NLP systems only rarely consider word-internal structure, beyond normalizing syntactic variation  \citep{Liu2012}.
Modern NLP methods such as language models employ subword tokenization, in which words are broken down  into smaller units.
This has an advantage of reducing vocabulary size, as well as learning shared representation between words with similar subwords.
The intent to improve such representation by making tokenization methods more linguistically sound has driven the invention of several  morphology segmentation methods, such as Morfessor \citep{gronroos-etal-2014-morfessor}. However, these have not been broadly adopted in modern LLMs as they do not scale well.

We hypothesize that current subword tokenization methods could lead to sub-optimal performance on language understanding tasks involving negation, because they do not correctly break words down morphologically. 
For instance, \Cref{tab:summary} demonstrates how different models employing different subword tokenization methods tokenize the word \textit{anticlinal}. Another known challenge which could affect models is the high false positive rate in detecting affixal negations \citep{blanco2011some}, for example misinterpreting \ex{de} in \ex{deserve} as a negative affix, where in practice the \ex{de} prefix derives from the Latin root \ex{deservire} and should not be interpreted as negating \ex{serve}. 

In this work, we analyze the impact of affixal negations on transformer-based language models, where two main tokenization methods are employed, namely: byte-pair encoding \citep{bpe-cage,sennrich-etal-2016-neural} and unigram language model \citep{kudo-2018-subword}. 
We 
consider
three research questions:
    \paragraph{RQ1: Are current subword tokenization methods able to preserve negative affixes?} We analyzed the performance of various subword tokenization methods used in modern LMs. We find that most 
    do not effectively produce the correct negative affixes. 
    
    \paragraph{RQ2: Are modern LMs aware of the presence of negation in affixal negations?} We design a negation prediction task to probe models' awareness of affixal negation. 
    We find that despite not performing well on the tokenization task, current LLMs can reliably infer the negated meaning of words with negative affixes. For this task, there is only a weak positive correlation between  tokenizer and classifier performance.
    
    \paragraph{RQ3: What are the impacts of affixal negation on downstream tasks?} 
    As negation and sentiment are closely related, we measure the impact on a downstream sentiment analysis task by looking at samples containing affixal negations from common datasets.
    Results show that models perform well on those samples, implying that the impact of affixal negation is minimal.
    However, there exists a bias in predicting negative sentiment for affixal negations.

\section{Related work}

There are two popular ways of constructing the vocabulary for LMs using subword tokenization methods: byte-pair encoding (``BPE'': \citet{sennrich-etal-2016-neural}) and unigram language model (``Unigram LM'': \citet{kudo-2018-subword}). 
BPE starts from a base character set, then merges those characters based on bigram frequency to form subword units (bottom-up), whereas unigram language models start from a large subword vocabulary, which is then reduced based on a regularization method (top-down).
There are multiple variants of BPE, differing in how the base vocabulary is represented and how the merging is done.
WordPiece \citep{wordpiece} uses characters to represent the base vocabulary, then selects pairs that maximize the likelihood of training data, 
Byte-level BPE uses bytes instead of Unicode to represent the base vocabulary; the merging is done based on the frequency count of bigrams. 
In contrast, the unigram language model  starts from a large base vocabulary and iteratively trims down tokens based on unigram LM perplexity until a target vocabulary size is reached.

\begin{figure*}[!t]
\centering
\begin{subfigure}[t]{0.45\textwidth}
\centering
\includegraphics[width=\columnwidth]{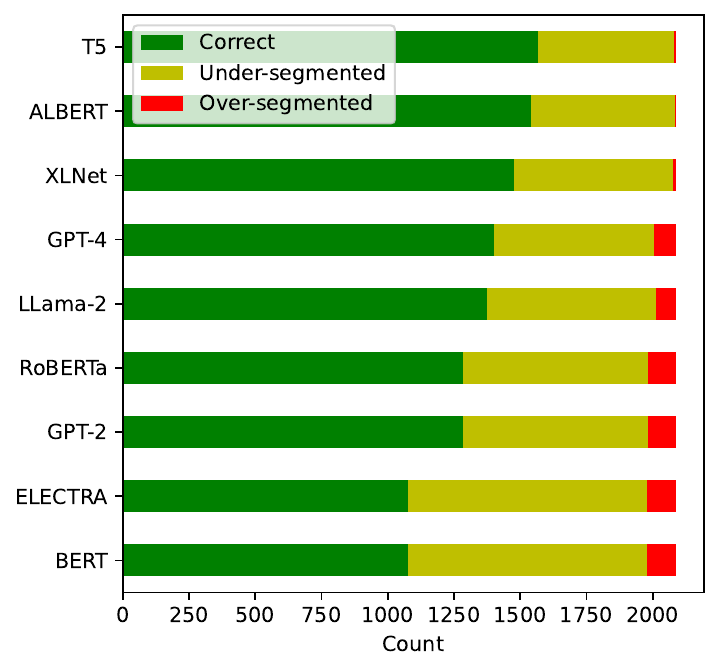}
\caption{NegMorph by models (ALBERT and XLNet use ULM, while the rest employ BPE)}
\label{fig:model-dist}
\end{subfigure}%
\begin{subfigure}[t]{0.45\textwidth}
\centering
\includegraphics[width=\columnwidth]{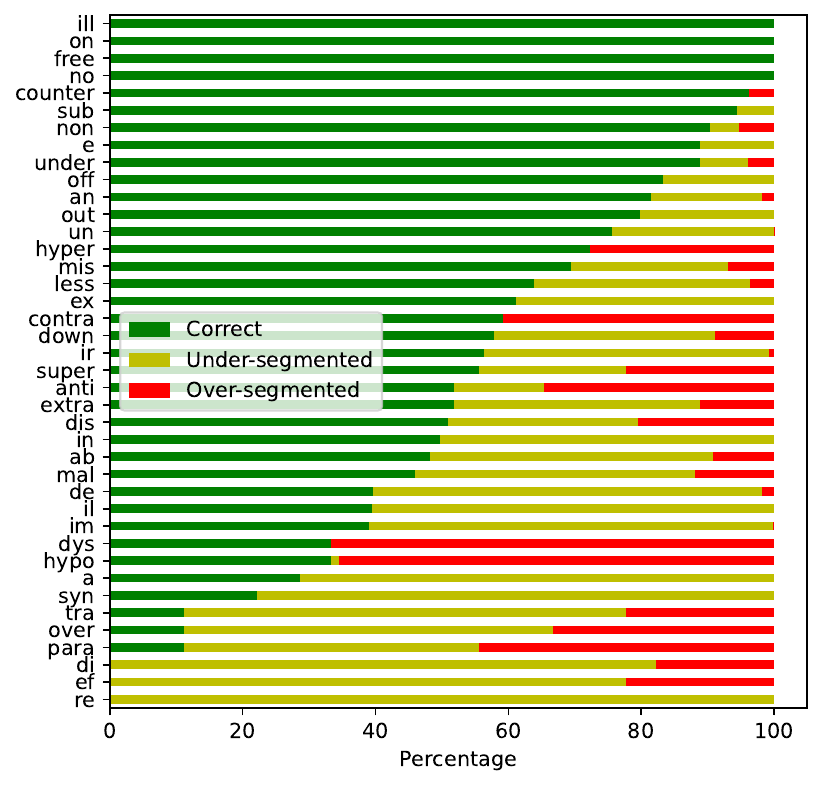}
\caption{NegMorph by affixes}
\label{fig:affix-dist}
\end{subfigure}
\caption{Negative affix-preserving segmentation performance on the set of affixal negations \citep{van-son-etal-2016-building}.}
\end{figure*}

Both methods assume that the input text uses spaces to separate words, which is not true for languages such as Chinese or Vietnamese. Therefore, a word segmentation step must be performed in advance. 
SentencePiece \citep{kudo-richardson-2018-sentencepiece} was introduced to solve this problem  by considering whitespace as part of words, essentially treating the whole input stream as the smallest unit to perform tokenization on.
Then, either BPE or unigram LM can be applied to construct the vocabulary. Regardless of method, they purely rely on statistical information and thus are not expected to produce morphologically-aligned subword tokens.

There have been efforts to build linguistically-sound word tokenization methods, most notably Morfessor and its variants \citep{gronroos-etal-2014-morfessor, gronroos-etal-2020-morfessor}.
Building morphology-aligned segmentation methods, especially in a multilingual setting, is an active line of research 
 through recent SIGMORPHON shared tasks \citep{batsuren-etal-2022-sigmorphon}. 
These methods outperform general  tokenizers in producing morphologically-aligned tokens,  but their benefit on downstream tasks is often negligible  \citep{domingo2019much,saleva-lignos-2021-effectiveness}. In this work, we examine if morphologically correct tokenization is important for LLMs to deal with negation.

BERT and its variants have been shown to be insensitive to negation \citep{kassner-schutze-2020-negated, ettinger-2020-bert}, affecting many downstream NLP tasks such as sentiment analysis, NLI, or QA \citep{hossain-etal-2020-analysis, hossain-etal-2022-analysis, ravichander-etal-2022-condaqa, truong-etal-2022-another}.
Compared to previous models, current LLMs have improved negation handling ability, but still struggle with some unconventional types of negation and linguistic constructions \citep{truong-etal-2023-language}. 
Here, we investigate the treatment of affixal negation in modern LMs, with the intuition that subword tokenization methods that don't appropriately reflect this morphology 
will lead to misinterpretation of their semantics.

\section{Experimental settings}
We focus our analysis particularly on how affixal negations are represented in modern LLMs, designing probing tasks to test their awareness of negation, and 
the effect on downstream tasks. All code for the experiments is available at \url{https://github.com/joey234/affixal-negation}.

\subsection{A lexicon of affixal negation}
\label{sec:dataset}
We use the lexicon created in \citet{van-son-etal-2016-building}. The dataset contains a list of affixal negation and their non-negated counterparts (e.g.\ \ex{unintended}--\ex{intended}).
For each affixal negation, the corresponding negative affix is also annotated.
In total, there are 2089 affixal negations, and 2055 non-negated words which are antonyms of the negations. 
These numbers are not equal because one word can have multiple corresponding negated counterparts, e.g.\ \ex{intrusive}--\ex{\{extrusive, unintrusive\}}.

\subsection{Tokenization methods}

For each tokenizer type (along with their variants), we consider the most representative models that use them, based on their popularity.
Although some models use the exact same tokenizer, it is worth investigating them as differences in training corpora can lead to differences in tokenization results.

\paragraph{BPE}
We consider models using different flavors of BPE. For WordPiece, we consider BERT \citep{devlin-etal-2019-bert}, ELECTRA \citep{clark2020electra}; for Byte-level BPE, we consider RoBERTa \citep{liu2019roberta}, and GPT-family models including GPT-2 \citep{radford2019language} and  GPT-4 \citep{openai2023gpt4}; and for SentencePiece, we examine Flan-T5 \citep{chung2022scaling} and LLaMA-2 \citep{touvron2023llama}.

\paragraph{Unigram LM}
Models using unigram LM tokenization methods considered in this work are always used in combination with SentencePiece: XLNet \citep{yang2019xlnet} and AlBERT \citep{lan2019albert}.

\subsection{Negative affix-preserving segmentation}

We consider a segmentation of an affixal negation to be ``affix preserving''  (\textbf{Correct}) only if the negative affix matches with one of the produced tokens (e.g.\ \ex{anticlimatic} $\rightarrow$ \ex{anti, clima, tic}). 
Otherwise, it is either \textbf{Under-segmented} if the negative affix is a substring of one of the produced tokens (e.g.\ \ex{anticlima, tic}), or \textbf{Over-segmented} (e.g.\ \ex{ant, i, clima, tic}).
Formally, given an affixal negation word $w$ having the negative affix $a$, if $w$ is tokenized into $T_k  = \{t_i,t_{i+1},...,t_n\}$ under tokenizer $k$ then we define $\text{NegMorph}_{k}(w)$ as follows:
\[\text{NegMorph}_{k}(w) = \begin{cases} 
\text{Correct} & \text{if $a \in T_k$.} \\
\parbox[t]{1.6cm}{Under-segmented} & \parbox[t]{2cm}{if $a$ is a substring of any $ t_i \in T_k$.} \\ 
\parbox[]{1.6cm}{Over-segmented} & \text{otherwise}
\end{cases}\]


\section{Findings}

\subsection{Current subword tokenization methods are not negative affix-preserving} 
As shown in \Cref{fig:model-dist}, T5 has the best performance in producing negative affix-preserving tokens, while for the remaining, models employing the unigram LM method outperform those using BPE.
This is in line with previous findings that the unigram LM produces subword units that align with morphology better than BPE \citep{bostrom-durrett-2020-byte}.
Moreover, models that employ SentencePiece (T5, ALBERT, XLNet, LLaMA-2) outperform those that don't (BERT, RoBERTa, GPT-2).
However, the best-performing models are only up to 75\% correct relative to NegMorph, with considerable room for improvement.
Most failed cases relate to under-segmentation.

\begin{figure}[t!]
    \centering
    \includegraphics[width = \columnwidth]{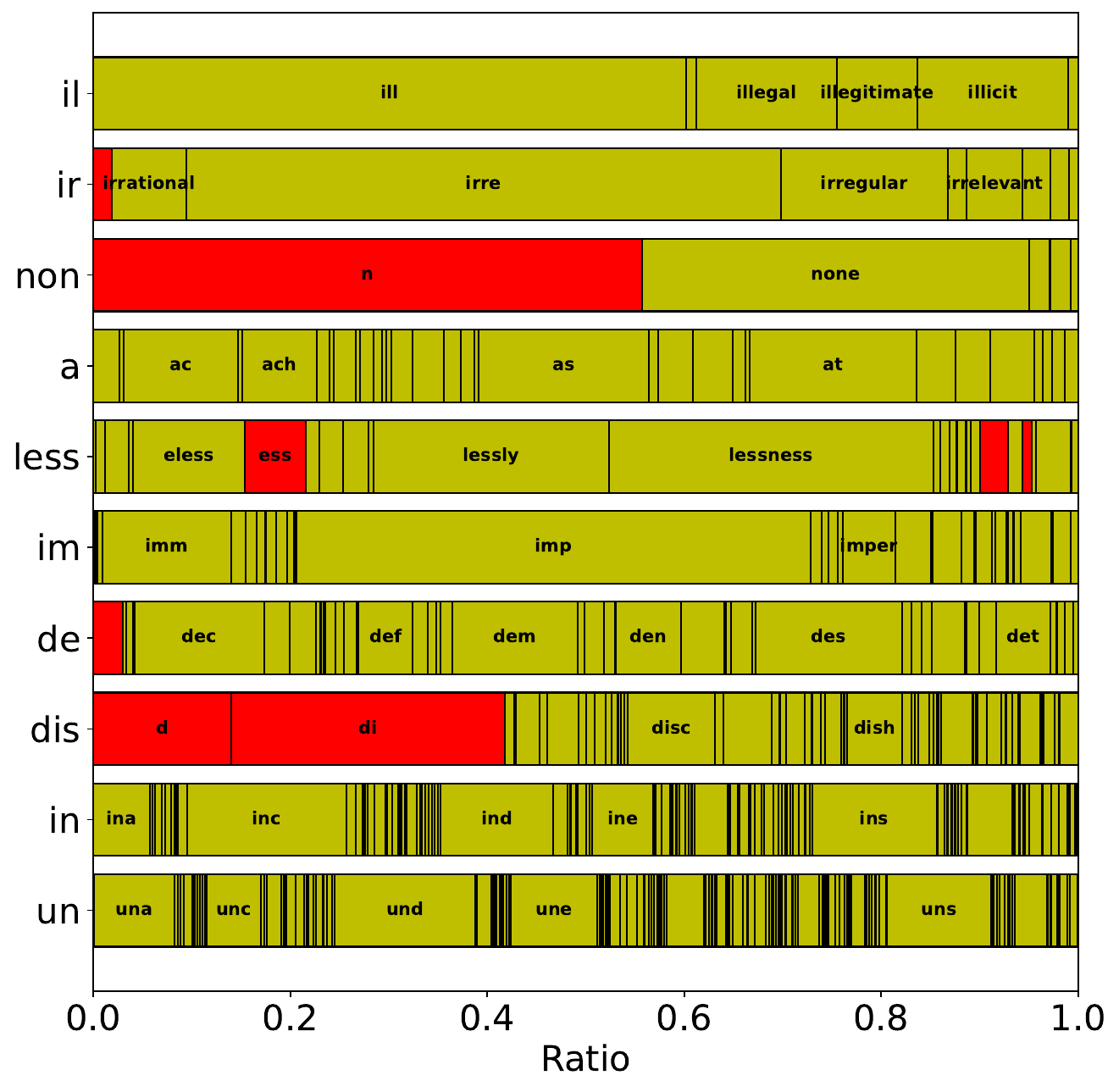}
    \caption{Top 10 most frequent affixes in the dataset and the distribution of tokens that they are wrongly tokenized into. Each substack denotes the percentage that the corresponding token is accounted for. {\color{yellow} Yellow bar} denotes Under-segmented, while {\color{red} Red bar} denotes Over-segmented.}
    \label{fig:negmorph-err}
\end{figure}

An analysis of what types of negative affix are hard to tokenize is provided in \Cref{fig:affix-dist}, and 
their most frequent incorrect tokenizations 
are shown in \Cref{fig:negmorph-err}.
Some common affixes that are incorrectly tokenized are \ex{il $\rightarrow$ ill} (\ex{illicit, illogical}), \ex{ir $\rightarrow$ irre} (\ex{irresolute, irreponsibly, irregular}), \ex{a $\rightarrow$ as} (\ex{asymmetric}), and \ex{a $\rightarrow$ at} (\ex{atypically}).
Overall, we see that some affixes can be wrongly tokenized in a wide range of ways (represented by the large number of substacks), showing that current tokenization methods are inefficient.
Overcoming this problem could result in embeddings that better encapsulate word morphology.

\begin{figure*}[t!]
    \centering
    \includegraphics[width =\linewidth]{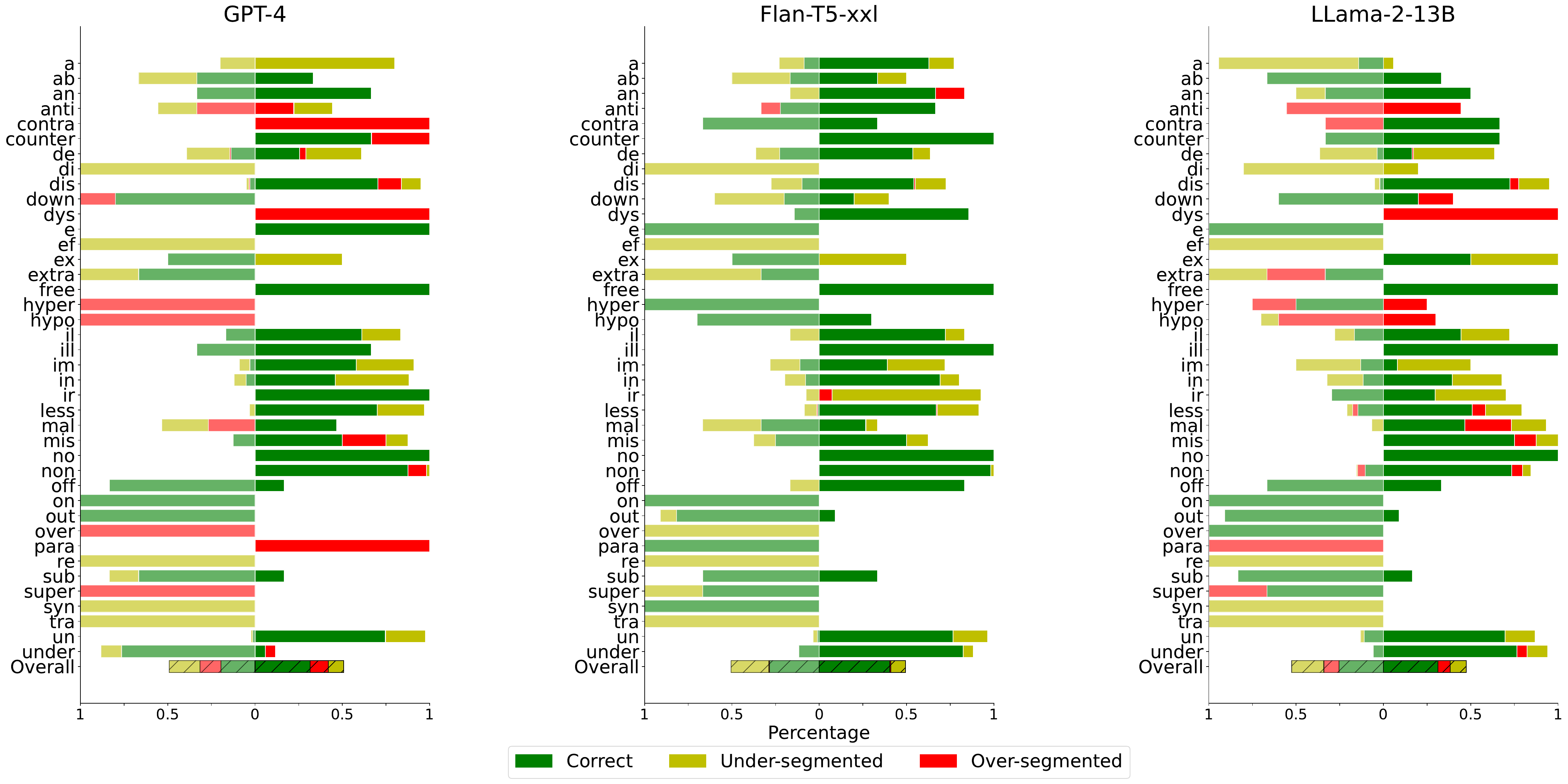}
    \caption{Ratio of correct/incorrect prediction on the Affix (fewshot) task, breakdown by affixes. The left greyed-out side of each subplot corresponds to wrong predictions.}
    \label{fig:result_breakdown}
\end{figure*}

\begin{figure}[t]
\centering
\begin{tcolorbox}
[colback=black!5!white,colframe=black!75!black,title=Affix (zero-shot)]
\scriptsize
\begin{verbatim}
The word {word} contains negation. True or 
False?
Answer:
\end{verbatim}
\end{tcolorbox}
\caption{Zero-shot prompt}
\label{fig:zero-shot-prompt}
\end{figure}

\begin{figure}[t]
\centering
\begin{tcolorbox}
[colback=black!5!white,colframe=black!75!black,title=Affix (few-shot)]
\scriptsize
\begin{verbatim}
A word contains negation if it has a negated 
meaning, usually expressed through a negative 
prefix (such as un, in) or suffix (such as 
less). 

The word decentralize contains negation. True 
or False?
Answer: True
Explanation: decentralize is created by 
prepending the root word centralize with the 
negative prefix de.

The word deserve contains negation. True or 
False?
Answer: False
Explanation: deserve just coincidentally starts 
with de.

The word {word} contains negation. True or 
False?
Answer:
\end{verbatim}
\end{tcolorbox}

\caption{Few-shot prompt}
\label{fig:few-shot-prompt}
\end{figure}

\subsection{Negative affixes signify negation, but word knowledge is essential}

We design a binary classification task on the lexicon described in \Cref{sec:dataset} to probe the ability of models to understand affixal negation, denoted \textit{Affix}.
First, for smaller models ($<$1B parameters), we conducted a fully fine-tuned setting with a 80/20 split and see that they achieve good results on the test set ($>$93\% accuracy), showing that models can learn the patterns of negative prefixes and suffixes with enough supervision.

For larger models, we evaluate three state-of-the-art LLMs in a zero- and few-shot manner.
The prompts are presented in \Cref{fig:zero-shot-prompt,fig:few-shot-prompt}, respectively.

\begin{figure}[!t]
    \centering
    \includegraphics[width =\columnwidth]{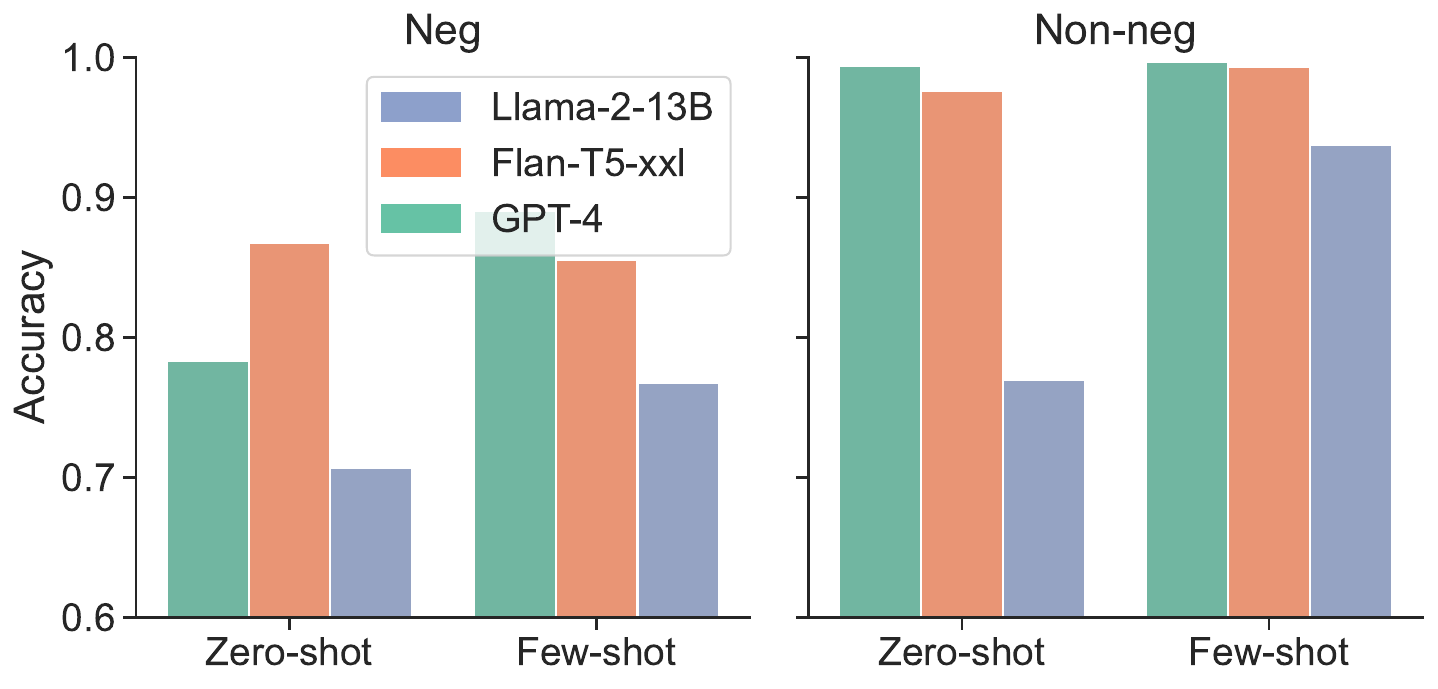}
    \caption{Zero- and Few-shot results on the affixal negation prediction task.}
    \label{fig:accuracy}
\end{figure}

For the few-shot prompt, we provide explicit instructions to explain what negation means in this context, as well as two demonstrating samples, to avoid ambiguity (such as confusion with negative sentiment). 

Results are summarized in \Cref{fig:accuracy} (full numerical results, including in the fine-tuned setting, are in \Cref{tab:accuracy}).
Overall, we find that the performance on Neg (the subset containing only affixal negation) is much lower compared to its Non-neg counterpart, where the best models achieve near-perfect performance.

For the zero-shot setting, surprisingly, Flan-T5 outperforms both LLaMA-2 and GPT-4, despite being the smallest in size.
After adding more explicit instructions and examples (Affix (few-shot)), we observe large increases in performance for GPT-4 and LLaMA-2, and little to no difference for Flan-T5.
For the non-negated subset, on the other hand, all models have near-perfect performance, with GPT-4 slightly outperforming Flan-T5.
LLaMA-2 performance for this task is much lower than the other two.

We further break down the results based on affixes. 
\Cref{fig:result_breakdown} illustrates the percentage of correct/incorrect prediction for each affix, 
divided by NegMorph categories.
Compared to the relatively high results for Neg in \Cref{tab:accuracy}, we have a clearer view on the actual performance of models.
On average, we see that models made errors equally as likely for all affixes (as shown by the last Overall bar, where the percentages of incorrect and correct predictions are roughly 50\%).
From the figure, we can also observe that the correct/incorrect prediction distribution is similar across models (especially between GPT-4 and Flan-T5), showing that they tend to make the same errors. 
In general, we see a larger portion of correct segmentation when models predict negation correctly, and more under-segmentation when models predict non-negation, while over-segmentation appears equally likely regardless of prediction. 
We also attempted to calculate the Pearson's coefficient between NegMorph and Accuracy on the Neg set but did not yield any statistically significant correlation.

\paragraph{Error analysis} 

We inspect the errors made by GPT-4 in the Affix (few-shot) setting to perform a qualitative analysis. We adopt the classification introduced in \citet{joshi2012affixal} and summarize in \Cref{tab:gpt4_error}.

\begin{table}[]
\footnotesize
    \centering
    \begin{tabular}{p{2.5cm} p{3cm} p{0.5cm}}
    \toprule
        \textbf{Error type}  & \textbf{Example} &  \textbf{Ratio}  \\
        \midrule
        Reversal of action &  \ex{divest, diverge, detach} & 0.213 \\
        
        Reversal of direction &  \ex{outdoors, descending, downstairs} & 0.204 \\

        Insufficiency & \ex{hypoglycemia, inferior, underpay} & 0.132  \\

        Positive sentiment & \ex{fearless, indispensable, unselfishly} & 0.081  \\

        Wrongness & \ex{infamy, malignant, misconstruction} & 0.047 \\
        
        Rare words & \ex{asyndetic, abactinal, syncategorem} & 0.047 \\
        
        Noise (annotation errors) & \ex{uncle, intense, increment} & 0.106 \\
        
        Other (normal affixal negation) & \ex{illicit, immortal, informality}  & 0.17 \\
    \bottomrule
    \end{tabular}
    \caption{Error analysis of the 235 errors made by GPT-4 on Affix (few-shot).}
    \label{tab:gpt4_error}
\end{table}

In total the model made 235 errors. Aside from the errors on normal affixal negation where the negative suffix can be replaced by \ex{not} without changing the word's meaning (\ex{uncritically, infinitely}), the bulk of the errors were caused by cases where the affix has more complex semantics. The most common source of errors is from affixes which show the reversal of action (e.g.\ \ex{diverge, detach}), or the reversal of direction (e.g.\ \ex{descending, downstairs}). In addition, a large proportion of errors come from affixes that negate reaching some normal or default state (e.g.\ \ex{hypoglycemia, inferior, underpay}).
Another interesting pattern is caused by words with positive sentiment (e.g.\  \ex{fearless, incredibly, infallibility}) showing that models are confused between negation and negative sentiment, likely due to over-exposure to sentiment analysis data. In addition, some errors are attributed to words where the negative affix has an additional sentiment of ``wrongness'' (e.g.\ \ex{malignant, misconstruction}).
The remaining errors are attributed to rare words and noise in annotation.

\begin{figure}[!t]
    \centering
    \includegraphics[width =\columnwidth]{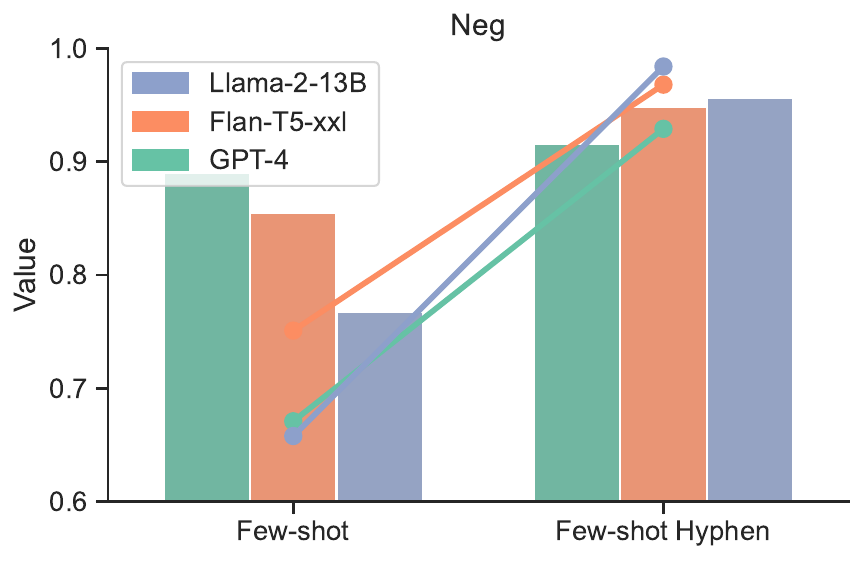}
    \caption{Results of Few-shot and Few-shot Hyphen on the affixal negation prediction task. Bars denote the accuracy on the prediction task , while Dots denote the Correct NegMorph scores for the segmentation task.}
    \label{fig:hyphen}
\end{figure}

\paragraph{Hyphenated words}
\label{sec:hyphen}
To make sure that the negative affix is not further broken down by tokenizers, we convert words into their ``hyphenated'' form (e.g.\ \ex{unintended} $\rightarrow$ \ex{un-intended}). 
From \Cref{fig:hyphen}, we see that this greatly increases the performance of different tokenizers on the NegMorph metric (by as much as 32\%).
Compared to the normal setting, the accuracy of all models also increases on the \textit{Affix} task, suggesting a positive correlation between NegMorph and Accuracy.
LLaMA-2 benefited the most from this setting, having the largest increases in both Accuracy and NegMorph.

\paragraph{Nonce words} 
\label{sec:nonce}

\begin{figure}
    \centering
    \includegraphics[width = \columnwidth]{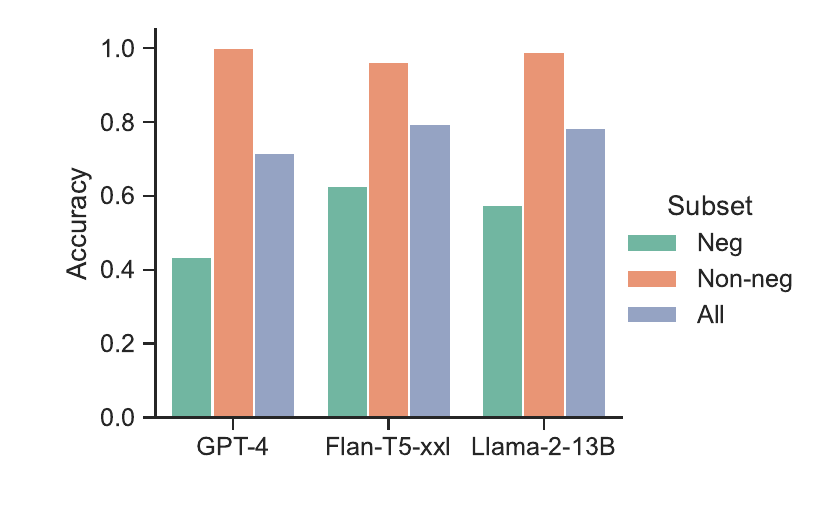}
    \caption{Accuracy on the affixal nonce words prediction task.}
    \label{fig:nonce-acc}
\end{figure}

Nonce words are words that look and sound like real words, but are created for a single-purpose use and not recognized as words within a language (e.g.\ \ex{roagly}). 
To measure the effect of negative affix on word semantics, we construct a list of ``affixal nonce words'' by prepending or appending negative affixes to a list of nonce words.
We collect a list of adjective nonce words from \citet{cremers2022interpreting} .
For affixes, we used the list of 40 negative affixes provided in \citet{van-son-etal-2016-building} and collected 40 non-negative affixes (e.g.\ \ex{auto-, bi-, -ism, -ful}).\footnote{{We collected the affixes from \scriptsize \url{https://litinfocus.com/120-root-words-prefixes-and-suffixes-pdf-list/}}}
For each nonce word, we prepend (or append) the affixes to form an ``affixal nonce word''.
In total, the set consists of 11 nonce words $\times$ 80 affixes = 880 samples, evenly distributed between negated (e.g.\ \ex{dis-roagly}) and non-negated (e.g.\ \ex{auto-roagly}).
We adopt the Affix (few-shot) prompt and  add an instruction to prevent models from refusing to answer the questions because of invalid words (full prompt in \Cref{sec:nonce_prompt}).
Similarly, we also report the results of two subsets of negative affixes (Neg) and non-negative affixes (Non-neg) in \Cref{fig:nonce-acc}.
For the Neg set, we find that the performance of all models is relatively low, despite them being able to correctly tokenize the negative affixes.
Whereas for the Non-neg set, performance is near-perfect for all models, similar to the previous Affix-Hyphen task.
Looking at the results, however, we found that most errors made by the models are when the negative affixes are ambiguous, i.e. their meaning depends on which words  they are attached to (e.g.\ \ex{a-, di-, ef-, para-, re-}).
This reveals an important insight that whether something is considered to be a negation should be judged with context (which is parametric knowledge about words in this case).

\begin{figure*}
    \centering
    \includegraphics[width =\textwidth]{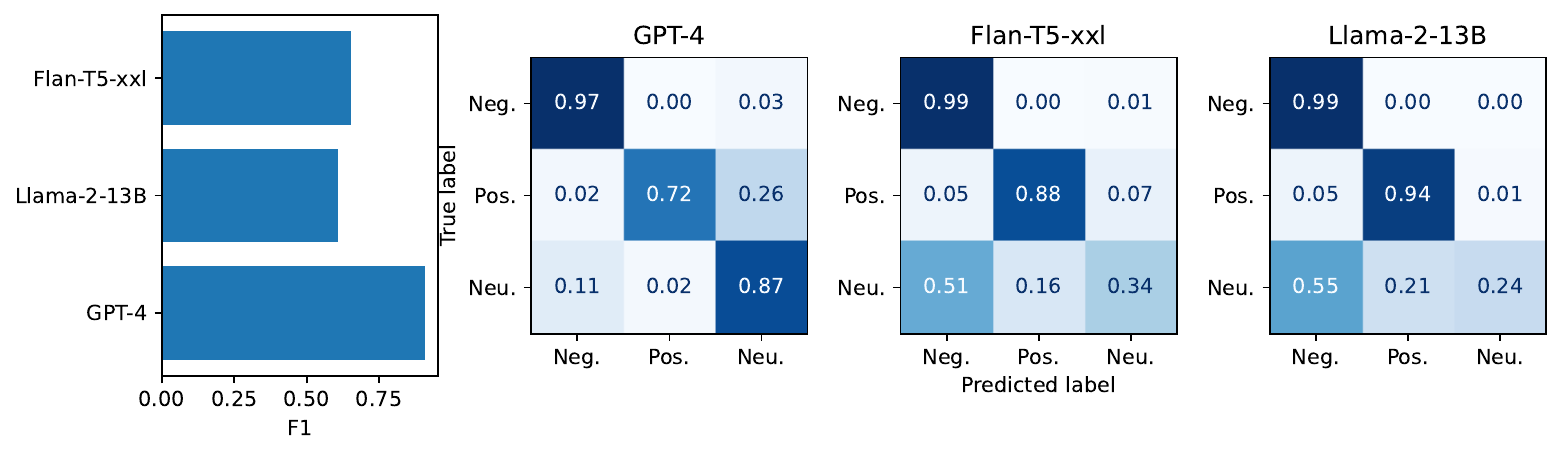}
    \caption{Performance on word-level sentiment task}
    \label{fig:word_sentiment}
\end{figure*}

\paragraph{Non-negated words with tokens homonymous with negative affixes}
To explore the false positive problem i.e., words coincidentally contain negative affixes,
we collect words from the Non-neg subset and WordNet \citep{miller1995wordnet} which do not have negated meaning, but have a negative prefix/suffix as the first/last subword token.
We tokenize WordNet using the T5 tokenizer and select all words that start/end with a negative prefix/suffix, then subtract all words in the list of affixal negations.
We manually go through the extracted list again to remove errors, resulting in a set of 330 words.\footnote{We didn't consider other models as this list of words would be different between models.}
Following the same affixal negation prediction task, we find that Flan-T5 has very good performance (0.958 accuracy), showing that it can synthesize information from all subword tokens instead of only relying on the negative affixes.
Most errors come from the \ex{uni-} cases, where the model tokenizes them as \ex{un-} (e.g.\ \ex{unidirectional, univalent}).


\subsection{Impact on downstream tasks}

One main drawback of our probing task is that the words lack context. 
Negation is a context-dependant concept, that is, what is considered negation can differ depending on the context of use. 
Investigating the impact of affixal negation in the context of downstream tasks is thus an essential component of this work.

\subsubsection{Sentiment analysis}
\label{sec:sentiment}
Previous work has shown that negation is a strong indicator of negative sentiment \citep{wiegand-etal-2010-survey}.
Furthermore, the fact that sentiment analysis is part of many NLP benchmarks could create a bias in models, leading to negation being conflated with negative sentiment, which is not always the case.
For instance, the word \ex{incredible} is constructed by prepending the root word \ex{credible} with the negative affix \ex{in-}, meaning ``not credible'' but used to express a positive meaning.
This inspired us to extend our analysis to a downstream sentiment analysis task.
We  evaluate the few-shot performance of LLMs in two settings of word- and sentence-level sentiment analysis (full prompts in \Cref{sec:sentiment-prompts}).

\paragraph{Word-level sentiment} We first use SentiWordNet 3.0 \citep{baccianella-etal-2010-sentiwordnet} to automatically assign a sentiment label for the lexicon of affixal negation described in \Cref{sec:dataset}.
After that, two graduate researchers went over the list to determine the final labels (positive, negative, or neutral).
In general, we find that GPT-4 outperforms Flan-T5 and LLaMA-2 on this word-level task.
As seen in \Cref{fig:word_sentiment}, all models have almost perfect performance at predicting negative words, but struggle with the other two classes.
In particular, we find Flan-T5 and LLaMA-2 overpredict Negative for Neutral words, while GPT-4 often mistakes Positive for Neutral.

\paragraph{Sentence-level sentiment}

\begin{figure}
    \centering
    \includegraphics[width = 0.7\columnwidth]{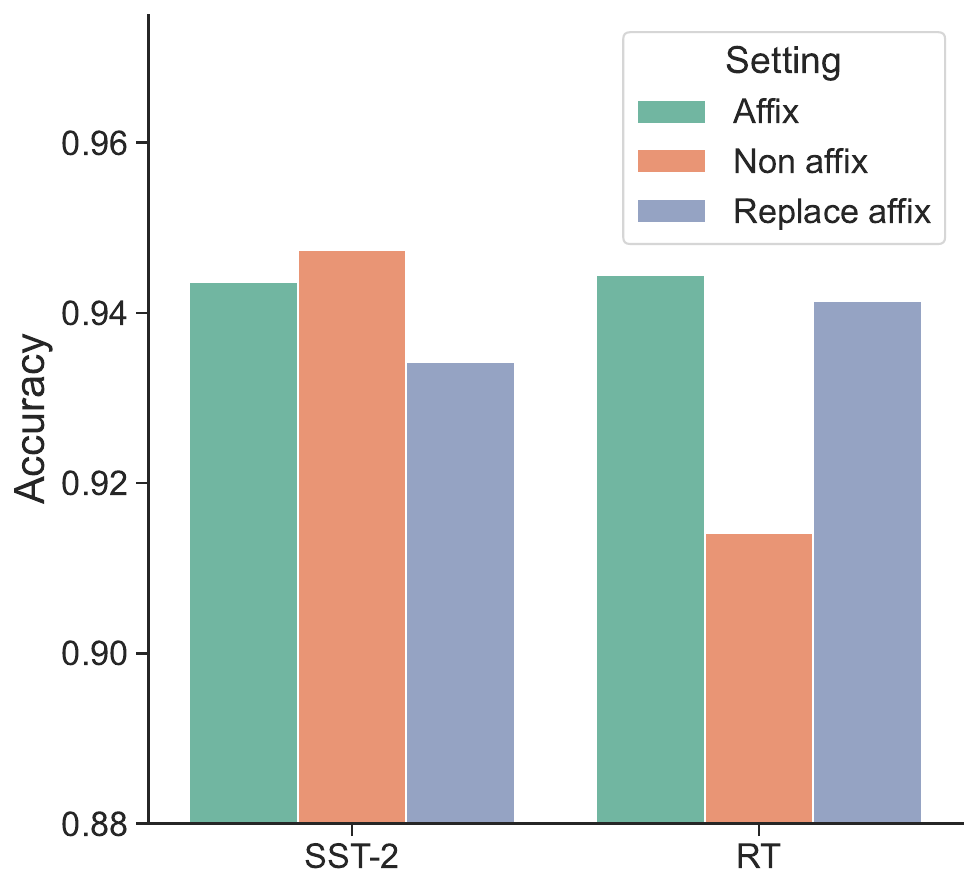}
    \caption{Accuracy on sentence-level sentiment analysis task. Results are averaged across 3 models.}
    \label{fig:sentiment-sentence}
\end{figure}

For this task, we look at common sentence-level sentiment analysis datasets including SST-2 \citep{socher-etal-2013-recursive}, and Rotten Tomatoes (RT) \citep{Pang+Lee:05a}.
One drawback of this evaluation is that samples tend to contain many sentiment signals, making it hard to gauge the effect of affixal negations.

\begin{figure}
    \centering
    \includegraphics[width = 0.6\columnwidth]{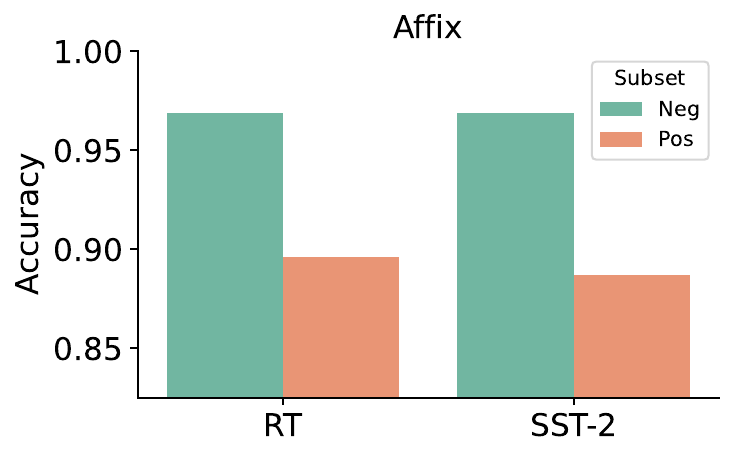}
    \caption{Accuracy of Neg/Pos class of the Affix set. Results are averaged across 3 models.}
    \label{fig:sentiment-sentence-class}
\end{figure}

We consider 3 settings: (1) Affix = only samples containing affixal negation; (2) Non affix = only samples without affixal negation; and (3) Replace affix = similar to Affix, but we replace all instances of affixal negations with equivalent syntactic negations, i.e. \ex{not} + word (\ex{uninteresting} $\rightarrow$ \ex{not interesting}). We present the results in \Cref{fig:sentiment-sentence}.
While it is true that replacing negative affixes with \ex{not} does not always result in a direct paraphrase, we argue that the change in meaning is minimal, and that samples will likely preserve the sentiment. 
Furthermore, we mostly find adjectives in the datasets rather than nouns, which ensures that the sentences are grammatically correct after replacement.
Overall, we can conclude that affixal negation is a strong signal to guide model prediction. We observe good performance for Affix in both datasets, where the accuracy are comparable to Non Affix in SST-2 and higher in RT. Attempting to replace affixal negations slightly decreases the performance of models in both datasets.
This suggests that affixal negation is actually a stronger sentiment cue compared to syntactic negation.  
We further report class-wise performance of the Affix set in \Cref{fig:sentiment-sentence-class}. Accuracy on samples with Negative sentiment is higher than Positive, once again showing that affixal negation is a strong cue for predicting negative sentiment.

%

\section{A look into token attribution}

We perform an interpretation analysis to attain  insights into what drives model predictions.
For this analysis, we use the Flan-T5-xxl model, as we could not obtain probabilities (logprobs) from GPT-4.
We calculate the attribution for each token corresponding to the predictions using the Integrated Gradient method \citep{sundararajan2017axiomatic}, with probability as the scoring function, implemented in \textit{Inseq} \citep{sarti-etal-2023-inseq}.
Overall, we observe high attribution scores from relevant tokens, such as the subword tokens of the target words, showing that models know where to pay attention to when performing inference.

\begin{figure}[!t]
\centering
\begin{subfigure}[t]{.4\linewidth}
  \includegraphics[width=\linewidth]{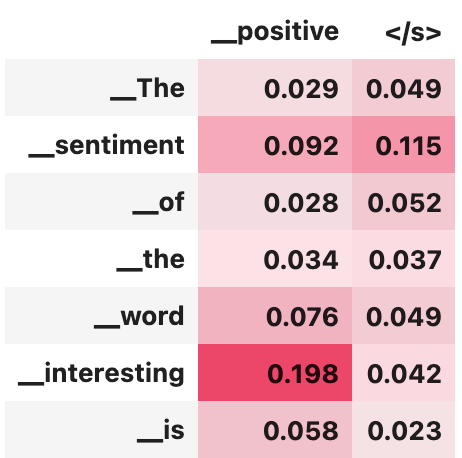}
\end{subfigure}
\begin{subfigure}[t]{.4\linewidth}
  \includegraphics[width=\linewidth]{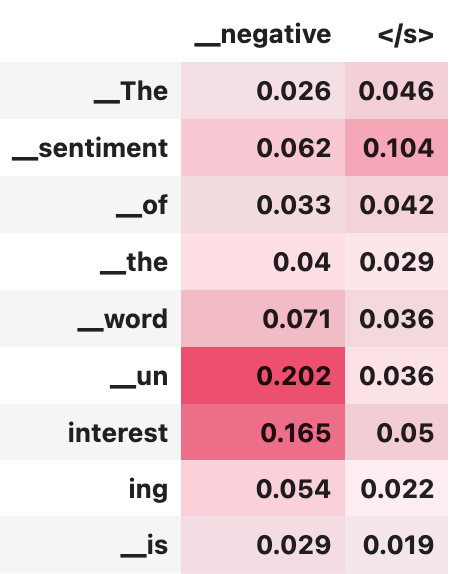}
\end{subfigure}

\begin{subfigure}[t]{.4\linewidth}
  \includegraphics[width=\linewidth]{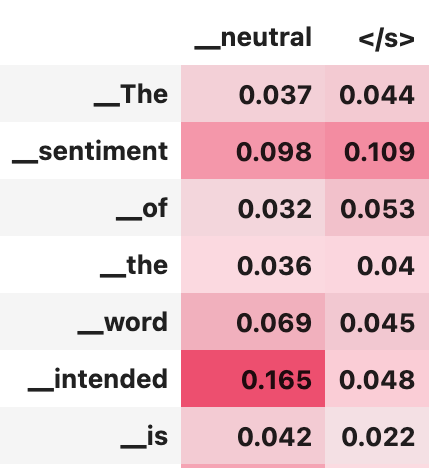}
\end{subfigure}
\begin{subfigure}[t]{.4\linewidth}
  \includegraphics[width=\linewidth]{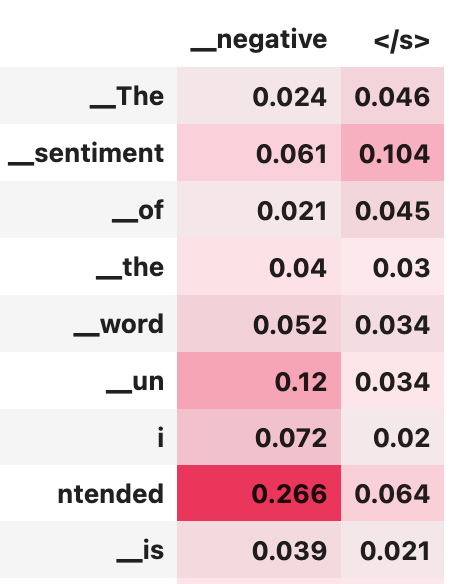}
\end{subfigure}

\caption{Token attribution of selected samples on the word-level sentiment prediction task. Only parts of the prompts are shown for clarity purposes.}
\label{fig:sentiment_saliency}
\end{figure}

\paragraph{Negative affixes have flipping sentiment effect}
In \Cref{sec:sentiment}, we see that models tend to overpredict negative sentiment on the list of affixal negations.
Through the saliency heatmap in \Cref{fig:sentiment_saliency}, we can see high token attributions for the negative affixes that change the sentiment  of the root words (either positive or neutral) into negative.
This is in line with previous findings that negation flips the polarity of sentiment \citep{tigges2023linear}.
This effect could be the main cause for the low performance on the Neutral class observed in our word-level sentiment analysis task (\Cref{sec:sentiment}).
When applied to the negation prediction task, however, we did not observe a similar effect and did not see any clear pattern for token attribution.

\begin{figure}[!htbp]
\centering
\begin{subfigure}[t]{.35\linewidth}
  \includegraphics[width=\linewidth]{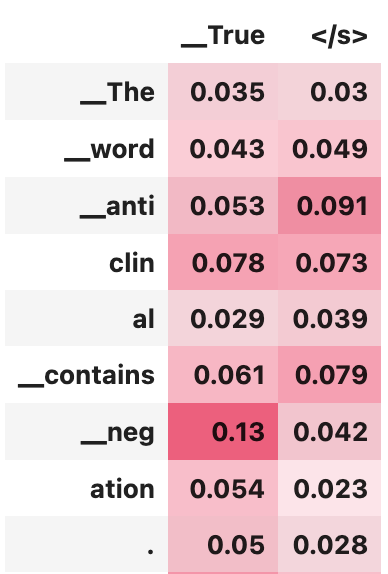}
\end{subfigure}
\begin{subfigure}[t]{.3\linewidth}
  \includegraphics[width=\linewidth]{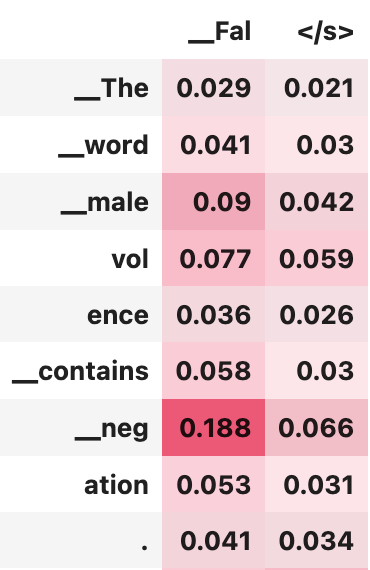}
\end{subfigure}
\begin{subfigure}[t]{.3\linewidth}
  \includegraphics[width=\linewidth]{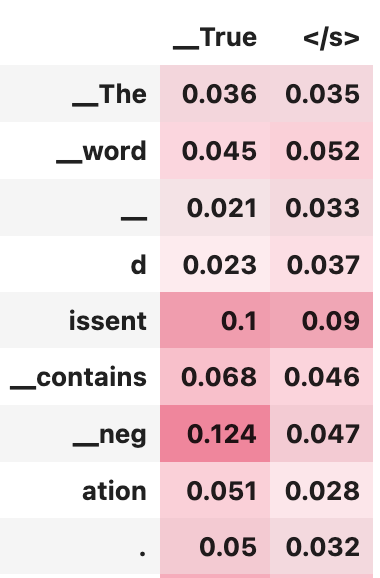}
\end{subfigure}
  
\caption{Token attribution of selected sample samples on the negation prediction task. The three subplots correspond to Correct, Under-segmented, and Over-segmented case respectively. Only parts of the prompts are shown for clarity purposes.}
\label{fig:negation_saliency}
\end{figure}

\paragraph{Correct tokenization is not essential for negation awareness} Through many experiments, we have shown that overall, correct tokenization leads to better awareness of models to the presence of negation. This effect, however is not significant. By comparing token attributions between 3 cases of NegMorph (\Cref{fig:negation_saliency}), we saw that models are able to combine information from relevant subword tokens corresponding to a word to make the correct inference. 

\section{Conclusion}
In this work, we conducted an in-depth analysis into how well modern LLMs handle affixal negation, a type of negation where morphology is essential to understanding word semantics. 
We have shown that there is significant room for improvement in  current tokenization methods in terms of producing negative affix-preserving tokens.
Despite that, the effect of morphologically incorrect tokenization on the ability of models to understand word meaning in downstream tasks, including sentiment analysis,  is minimal.
Regardless, designing better subword tokenization methods may have many immediate benefits such as reducing vocabulary size, learning better  word representations, and improving model interpretability.

\section{Limitations}
\label{sec:limitation}

\paragraph{Prompting} As this work involves experiments with LLMs, there is always a possibility that the prompts we used are not optimal (and also, the problem of reproducibility). We attempted to reuse prompt templates from existing work where possible, and strove to design prompts that are intuitive and specific otherwise.

\paragraph{Multilinguality} Morphology is a language-dependent problem. We recognize that the lack of investigation in other languages other than English is a drawback of this work.

\paragraph{Broader impact} Given that our focus is on presenting and analysing the problem of poor treatment of affixal negation in LLMs, we did not propose any immediate solutions to improve the status quo.  The finding on the impact on downstream tasks  could be limited by the lack of samples (both in size and meaningful patterns) in the test data.

\section*{Acknowledgements}
We would like to thank Khuyagbaatar Batsuren for the talk on morphological segmentation at The University of Melbourne that inspired this work, and Daniel Beck and  Simon Šuster for discussions that helped refine our experiments. 
We also thank the anonymous reviewers and area chairs of ARR December cycle and NAACL for an engaging and constructive review period. 
This research was conducted by the Australian Research Council Training Centre in Cognitive Computing for Medical Technologies (project number ICI70200030) and funded by the Australian Government.
This research was undertaken using the LIEF HPC-GPGPU Facility hosted at The University of Melbourne, established with the assistance of LIEF Grant LE170100200.

\bibliography{anthology,custom}

\begin{thebibliography}{41}
\expandafter\ifx\csname natexlab\endcsname\relax\def\natexlab#1{#1}\fi

\bibitem[{Baccianella et~al.(2010)Baccianella, Esuli, and Sebastiani}]{baccianella-etal-2010-sentiwordnet}
Stefano Baccianella, Andrea Esuli, and Fabrizio Sebastiani. 2010.
\newblock \href {http://www.lrec-conf.org/proceedings/lrec2010/pdf/769_Paper.pdf} {{S}enti{W}ord{N}et 3.0: An enhanced lexical resource for sentiment analysis and opinion mining}.
\newblock In \emph{Proceedings of the Seventh International Conference on Language Resources and Evaluation ({LREC}'10)}, Valletta, Malta. European Language Resources Association (ELRA).

\bibitem[{Batsuren et~al.(2022)Batsuren, Bella, Arora, Martinovic, Gorman, {\v{Z}}abokrtsk{\'y}, Ganbold, Dohnalov{\'a}, {\v{S}}ev{\v{c}}{\'\i}kov{\'a}, Pelegrinov{\'a}, Giunchiglia, Cotterell, and Vylomova}]{batsuren-etal-2022-sigmorphon}
Khuyagbaatar Batsuren, G{\'a}bor Bella, Aryaman Arora, Viktor Martinovic, Kyle Gorman, Zden{\v{e}}k {\v{Z}}abokrtsk{\'y}, Amarsanaa Ganbold, {\v{S}}{\'a}rka Dohnalov{\'a}, Magda {\v{S}}ev{\v{c}}{\'\i}kov{\'a}, Kate{\v{r}}ina Pelegrinov{\'a}, Fausto Giunchiglia, Ryan Cotterell, and Ekaterina Vylomova. 2022.
\newblock \href {https://doi.org/10.18653/v1/2022.sigmorphon-1.11} {The {SIGMORPHON} 2022 shared task on morpheme segmentation}.
\newblock In \emph{Proceedings of the 19th SIGMORPHON Workshop on Computational Research in Phonetics, Phonology, and Morphology}, pages 103--116, Seattle, Washington. Association for Computational Linguistics.

\bibitem[{Blanco and Moldovan(2011)}]{blanco2011some}
Eduardo Blanco and Dan Moldovan. 2011.
\newblock Some issues on detecting negation from text.
\newblock In \emph{Twenty-Fourth International FLAIRS Conference}.

\bibitem[{Bostrom and Durrett(2020)}]{bostrom-durrett-2020-byte}
Kaj Bostrom and Greg Durrett. 2020.
\newblock \href {https://doi.org/10.18653/v1/2020.findings-emnlp.414} {Byte pair encoding is suboptimal for language model pretraining}.
\newblock In \emph{Findings of the Association for Computational Linguistics: EMNLP 2020}, pages 4617--4624, Online. Association for Computational Linguistics.

\bibitem[{Chung et~al.(2022)Chung, Hou, Longpre, Zoph, Tay, Fedus, Li, Wang, Dehghani, Brahma, Webson, Gu, Dai, Suzgun, Chen, Chowdhery, Castro-Ros, Pellat, Robinson, Valter, Narang, Mishra, Yu, Zhao, Huang, Dai, Yu, Petrov, Chi, Dean, Devlin, Roberts, Zhou, Le, and Wei}]{chung2022scaling}
Hyung~Won Chung, Le~Hou, Shayne Longpre, Barret Zoph, Yi~Tay, William Fedus, Yunxuan Li, Xuezhi Wang, Mostafa Dehghani, Siddhartha Brahma, Albert Webson, Shixiang~Shane Gu, Zhuyun Dai, Mirac Suzgun, Xinyun Chen, Aakanksha Chowdhery, Alex Castro-Ros, Marie Pellat, Kevin Robinson, Dasha Valter, Sharan Narang, Gaurav Mishra, Adams Yu, Vincent Zhao, Yanping Huang, Andrew Dai, Hongkun Yu, Slav Petrov, Ed~H. Chi, Jeff Dean, Jacob Devlin, Adam Roberts, Denny Zhou, Quoc~V. Le, and Jason Wei. 2022.
\newblock \href {https://arxiv.org/abs/2210.11416} {Scaling instruction-finetuned language models}.
\newblock \emph{ArXiv preprint}, abs/2210.11416.

\bibitem[{Clark et~al.(2020)Clark, Luong, Le, and Manning}]{clark2020electra}
Kevin Clark, Minh{-}Thang Luong, Quoc~V. Le, and Christopher~D. Manning. 2020.
\newblock \href {https://openreview.net/forum?id=r1xMH1BtvB} {{ELECTRA:} pre-training text encoders as discriminators rather than generators}.
\newblock In \emph{8th International Conference on Learning Representations ({ICLR} 2020)}.

\bibitem[{Cremers(2022)}]{cremers2022interpreting}
Alexandre Cremers. 2022.
\newblock Interpreting gradable adjectives: rational reasoning or simple heuristics?
\newblock \emph{Empirical Issues in Syntax and Semantics}, 14:31--61.

\bibitem[{Devlin et~al.(2019)Devlin, Chang, Lee, and Toutanova}]{devlin-etal-2019-bert}
Jacob Devlin, Ming-Wei Chang, Kenton Lee, and Kristina Toutanova. 2019.
\newblock \href {https://doi.org/10.18653/v1/N19-1423} {{BERT}: Pre-training of deep bidirectional transformers for language understanding}.
\newblock In \emph{Proceedings of the 2019 Conference of the North {A}merican Chapter of the Association for Computational Linguistics: Human Language Technologies, Volume 1 (Long and Short Papers)}, pages 4171--4186, Minneapolis, Minnesota. Association for Computational Linguistics.

\bibitem[{Domingo et~al.(2019)Domingo, Garc{\'\i}a-Mart{\'\i}nez, Helle, Casacuberta, and Herranz}]{domingo2019much}
Miguel Domingo, Mercedes Garc{\'\i}a-Mart{\'\i}nez, Alexandre Helle, Francisco Casacuberta, and Manuel Herranz. 2019.
\newblock How much does tokenization affect neural machine translation?
\newblock In \emph{International Conference on Computational Linguistics and Intelligent Text Processing}, pages 545--554. Springer.

\bibitem[{Ettinger(2020)}]{ettinger-2020-bert}
Allyson Ettinger. 2020.
\newblock \href {https://doi.org/10.1162/tacl_a_00298} {What {BERT} is not: Lessons from a new suite of psycholinguistic diagnostics for language models}.
\newblock \emph{Transactions of the Association for Computational Linguistics}, 8:34--48.

\bibitem[{Gage(1994)}]{bpe-cage}
Philip Gage. 1994.
\newblock A new algorithm for data compression.
\newblock \emph{C Users Journal}, 12(2):23–38.

\bibitem[{Gr{\"o}nroos et~al.(2020)Gr{\"o}nroos, Virpioja, and Kurimo}]{gronroos-etal-2020-morfessor}
Stig-Arne Gr{\"o}nroos, Sami Virpioja, and Mikko Kurimo. 2020.
\newblock \href {https://aclanthology.org/2020.lrec-1.486} {{M}orfessor {EM}+{P}rune: Improved subword segmentation with expectation maximization and pruning}.
\newblock In \emph{Proceedings of the Twelfth Language Resources and Evaluation Conference}, pages 3944--3953, Marseille, France. European Language Resources Association.

\bibitem[{Gr{\"o}nroos et~al.(2014)Gr{\"o}nroos, Virpioja, Smit, and Kurimo}]{gronroos-etal-2014-morfessor}
Stig-Arne Gr{\"o}nroos, Sami Virpioja, Peter Smit, and Mikko Kurimo. 2014.
\newblock \href {https://aclanthology.org/C14-1111} {{M}orfessor {F}lat{C}at: An {HMM}-based method for unsupervised and semi-supervised learning of morphology}.
\newblock In \emph{Proceedings of {COLING} 2014, the 25th International Conference on Computational Linguistics: Technical Papers}, pages 1177--1185, Dublin, Ireland. Dublin City University and Association for Computational Linguistics.

\bibitem[{Hossain et~al.(2022)Hossain, Chinnappa, and Blanco}]{hossain-etal-2022-analysis}
Md~Mosharaf Hossain, Dhivya Chinnappa, and Eduardo Blanco. 2022.
\newblock \href {https://doi.org/10.18653/v1/2022.acl-short.81} {An analysis of negation in natural language understanding corpora}.
\newblock In \emph{Proceedings of the 60th Annual Meeting of the Association for Computational Linguistics (Volume 2: Short Papers)}, pages 716--723, Dublin, Ireland. Association for Computational Linguistics.

\bibitem[{Hossain et~al.(2020)Hossain, Kovatchev, Dutta, Kao, Wei, and Blanco}]{hossain-etal-2020-analysis}
Md~Mosharaf Hossain, Venelin Kovatchev, Pranoy Dutta, Tiffany Kao, Elizabeth Wei, and Eduardo Blanco. 2020.
\newblock \href {https://doi.org/10.18653/v1/2020.emnlp-main.732} {An analysis of natural language inference benchmarks through the lens of negation}.
\newblock In \emph{Proceedings of the 2020 Conference on Empirical Methods in Natural Language Processing (EMNLP)}, pages 9106--9118, Online. Association for Computational Linguistics.

\bibitem[{Joshi(2012)}]{joshi2012affixal}
Shrikant Joshi. 2012.
\newblock Affixal negation--direct, indirect and their subtypes 1.
\newblock \emph{Syntaxe \& s{\'e}mantique}, (1):49--63.

\bibitem[{Kassner and Sch{\"u}tze(2020)}]{kassner-schutze-2020-negated}
Nora Kassner and Hinrich Sch{\"u}tze. 2020.
\newblock \href {https://doi.org/10.18653/v1/2020.acl-main.698} {Negated and misprimed probes for pretrained language models: Birds can talk, but cannot fly}.
\newblock In \emph{Proceedings of the 58th Annual Meeting of the Association for Computational Linguistics}, pages 7811--7818, Online. Association for Computational Linguistics.

\bibitem[{Kudo(2018)}]{kudo-2018-subword}
Taku Kudo. 2018.
\newblock \href {https://doi.org/10.18653/v1/P18-1007} {Subword regularization: Improving neural network translation models with multiple subword candidates}.
\newblock In \emph{Proceedings of the 56th Annual Meeting of the Association for Computational Linguistics (Volume 1: Long Papers)}, pages 66--75, Melbourne, Australia. Association for Computational Linguistics.

\bibitem[{Kudo and Richardson(2018)}]{kudo-richardson-2018-sentencepiece}
Taku Kudo and John Richardson. 2018.
\newblock \href {https://doi.org/10.18653/v1/D18-2012} {{S}entence{P}iece: A simple and language independent subword tokenizer and detokenizer for neural text processing}.
\newblock In \emph{Proceedings of the 2018 Conference on Empirical Methods in Natural Language Processing: System Demonstrations}, pages 66--71, Brussels, Belgium. Association for Computational Linguistics.

\bibitem[{Lan et~al.(2020)Lan, Chen, Goodman, Gimpel, Sharma, and Soricut}]{lan2019albert}
Zhenzhong Lan, Mingda Chen, Sebastian Goodman, Kevin Gimpel, Piyush Sharma, and Radu Soricut. 2020.
\newblock \href {https://openreview.net/forum?id=H1eA7AEtvS} {{ALBERT:} {A} lite {BERT} for self-supervised learning of language representations}.
\newblock In \emph{8th International Conference on Learning Representations ({ICLR} 2020)}.

\bibitem[{Liu et~al.(2012)Liu, Christiansen, Baumgartner, and Verspoor}]{Liu2012}
Haibin Liu, Tom Christiansen, William~A Baumgartner, and Karin Verspoor. 2012.
\newblock \href {https://doi.org/10.1186/2041-1480-3-3} {Biolemmatizer: a lemmatization tool for morphological processing of biomedical text.}
\newblock \emph{Journal of Biomedical Semantics}, 3:3.

\bibitem[{Liu et~al.(2019)Liu, Ott, Goyal, Du, Joshi, Chen, Levy, Lewis, Zettlemoyer, and Stoyanov}]{liu2019roberta}
Yinhan Liu, Myle Ott, Naman Goyal, Jingfei Du, Mandar Joshi, Danqi Chen, Omer Levy, Mike Lewis, Luke Zettlemoyer, and Veselin Stoyanov. 2019.
\newblock \href {https://arxiv.org/abs/1907.11692} {{RoBERTa}: A robustly optimized {BERT} pretraining approach}.
\newblock \emph{ArXiv preprint}, abs/1907.11692.

\bibitem[{Miller(1995)}]{miller1995wordnet}
George~A Miller. 1995.
\newblock {WordNet}: a lexical database for {English}.
\newblock \emph{Communications of the ACM}, 38(11):39--41.

\bibitem[{OpenAI(2023)}]{openai2023gpt4}
OpenAI. 2023.
\newblock \href {http://arxiv.org/abs/2303.08774} {{GPT-4} technical report}.

\bibitem[{Pang and Lee(2005)}]{Pang+Lee:05a}
Bo~Pang and Lillian Lee. 2005.
\newblock Seeing stars: Exploiting class relationships for sentiment categorization with respect to rating scales.
\newblock In \emph{Proceedings of the ACL}.

\bibitem[{Radford et~al.(2019)Radford, Wu, Child, Luan, Amodei, and Sutskever}]{radford2019language}
Alec Radford, Jeff Wu, Rewon Child, David Luan, Dario Amodei, and Ilya Sutskever. 2019.
\newblock Language models are unsupervised multitask learners.
\newblock OpenAI blog.

\bibitem[{Ravichander et~al.(2022)Ravichander, Gardner, and Marasovic}]{ravichander-etal-2022-condaqa}
Abhilasha Ravichander, Matt Gardner, and Ana Marasovic. 2022.
\newblock \href {https://doi.org/10.18653/v1/2022.emnlp-main.598} {{CONDAQA}: A contrastive reading comprehension dataset for reasoning about negation}.
\newblock In \emph{Proceedings of the 2022 Conference on Empirical Methods in Natural Language Processing}, pages 8729--8755, Abu Dhabi, United Arab Emirates. Association for Computational Linguistics.

\bibitem[{Saleva and Lignos(2021)}]{saleva-lignos-2021-effectiveness}
Jonne Saleva and Constantine Lignos. 2021.
\newblock \href {https://doi.org/10.18653/v1/2021.eacl-srw.22} {The effectiveness of morphology-aware segmentation in low-resource neural machine translation}.
\newblock In \emph{Proceedings of the 16th Conference of the European Chapter of the Association for Computational Linguistics: Student Research Workshop}, pages 164--174, Online. Association for Computational Linguistics.

\bibitem[{Sarti et~al.(2023)Sarti, Feldhus, Sickert, and van~der Wal}]{sarti-etal-2023-inseq}
Gabriele Sarti, Nils Feldhus, Ludwig Sickert, and Oskar van~der Wal. 2023.
\newblock \href {https://doi.org/10.18653/v1/2023.acl-demo.40} {Inseq: An interpretability toolkit for sequence generation models}.
\newblock In \emph{Proceedings of the 61st Annual Meeting of the Association for Computational Linguistics (Volume 3: System Demonstrations)}, pages 421--435, Toronto, Canada. Association for Computational Linguistics.

\bibitem[{Schuster and Nakajima(2012)}]{wordpiece}
Mike Schuster and Kaisuke Nakajima. 2012.
\newblock Japanese and korean voice search.
\newblock In \emph{International Conference on Acoustics, Speech and Signal Processing}, pages 5149--5152.

\bibitem[{Sennrich et~al.(2016)Sennrich, Haddow, and Birch}]{sennrich-etal-2016-neural}
Rico Sennrich, Barry Haddow, and Alexandra Birch. 2016.
\newblock \href {https://doi.org/10.18653/v1/P16-1162} {Neural machine translation of rare words with subword units}.
\newblock In \emph{Proceedings of the 54th Annual Meeting of the Association for Computational Linguistics (Volume 1: Long Papers)}, pages 1715--1725, Berlin, Germany. Association for Computational Linguistics.

\bibitem[{She et~al.(2023)She, Potts, Bowman, and Geiger}]{she-etal-2023-scone}
Jingyuan~S. She, Christopher Potts, Samuel~R. Bowman, and Atticus Geiger. 2023.
\newblock \href {https://doi.org/10.18653/v1/2023.acl-short.154} {{S}co{N}e: Benchmarking negation reasoning in language models with fine-tuning and in-context learning}.
\newblock In \emph{Proceedings of the 61st Annual Meeting of the Association for Computational Linguistics (Volume 2: Short Papers)}, pages 1803--1821, Toronto, Canada. Association for Computational Linguistics.

\bibitem[{Socher et~al.(2013)Socher, Perelygin, Wu, Chuang, Manning, Ng, and Potts}]{socher-etal-2013-recursive}
Richard Socher, Alex Perelygin, Jean Wu, Jason Chuang, Christopher~D. Manning, Andrew Ng, and Christopher Potts. 2013.
\newblock \href {https://aclanthology.org/D13-1170} {Recursive deep models for semantic compositionality over a sentiment treebank}.
\newblock In \emph{Proceedings of the 2013 Conference on Empirical Methods in Natural Language Processing}, pages 1631--1642, Seattle, Washington, USA. Association for Computational Linguistics.

\bibitem[{Sundararajan et~al.(2017)Sundararajan, Taly, and Yan}]{sundararajan2017axiomatic}
Mukund Sundararajan, Ankur Taly, and Qiqi Yan. 2017.
\newblock Axiomatic attribution for deep networks.
\newblock In \emph{International conference on machine learning}, pages 3319--3328. PMLR.

\bibitem[{Tigges et~al.(2023)Tigges, Hollinsworth, Geiger, and Nanda}]{tigges2023linear}
Curt Tigges, Oskar~John Hollinsworth, Atticus Geiger, and Neel Nanda. 2023.
\newblock \href {http://arxiv.org/abs/2310.15154} {Linear representations of sentiment in large language models}.

\bibitem[{Touvron et~al.(2023)Touvron, Martin, Stone, Albert, Almahairi, Babaei, Bashlykov, Batra, Bhargava, Bhosale, Bikel, Blecher, Ferrer, Chen, Cucurull, Esiobu, Fernandes, Fu, Fu, Fuller, Gao, Goswami, Goyal, Hartshorn, Hosseini, Hou, Inan, Kardas, Kerkez, Khabsa, Kloumann, Korenev, Koura, Lachaux, Lavril, Lee, Liskovich, Lu, Mao, Martinet, Mihaylov, Mishra, Molybog, Nie, Poulton, Reizenstein, Rungta, Saladi, Schelten, Silva, Smith, Subramanian, Tan, Tang, Taylor, Williams, Kuan, Xu, Yan, Zarov, Zhang, Fan, Kambadur, Narang, Rodriguez, Stojnic, Edunov, and Scialom}]{touvron2023llama}
Hugo Touvron, Louis Martin, Kevin Stone, Peter Albert, Amjad Almahairi, Yasmine Babaei, Nikolay Bashlykov, Soumya Batra, Prajjwal Bhargava, Shruti Bhosale, Dan Bikel, Lukas Blecher, Cristian~Canton Ferrer, Moya Chen, Guillem Cucurull, David Esiobu, Jude Fernandes, Jeremy Fu, Wenyin Fu, Brian Fuller, Cynthia Gao, Vedanuj Goswami, Naman Goyal, Anthony Hartshorn, Saghar Hosseini, Rui Hou, Hakan Inan, Marcin Kardas, Viktor Kerkez, Madian Khabsa, Isabel Kloumann, Artem Korenev, Punit~Singh Koura, Marie-Anne Lachaux, Thibaut Lavril, Jenya Lee, Diana Liskovich, Yinghai Lu, Yuning Mao, Xavier Martinet, Todor Mihaylov, Pushkar Mishra, Igor Molybog, Yixin Nie, Andrew Poulton, Jeremy Reizenstein, Rashi Rungta, Kalyan Saladi, Alan Schelten, Ruan Silva, Eric~Michael Smith, Ranjan Subramanian, Xiaoqing~Ellen Tan, Binh Tang, Ross Taylor, Adina Williams, Jian~Xiang Kuan, Puxin Xu, Zheng Yan, Iliyan Zarov, Yuchen Zhang, Angela Fan, Melanie Kambadur, Sharan Narang, Aurelien Rodriguez, Robert Stojnic, Sergey Edunov, and Thomas
  Scialom. 2023.
\newblock {LLaMA} 2: Open foundation and fine-tuned chat models.
\newblock \emph{arXiv preprint arXiv:2307.09288}.

\bibitem[{Truong et~al.(2023)Truong, Baldwin, Verspoor, and Cohn}]{truong-etal-2023-language}
Thinh~Hung Truong, Timothy Baldwin, Karin Verspoor, and Trevor Cohn. 2023.
\newblock \href {https://doi.org/10.18653/v1/2023.starsem-1.10} {Language models are not naysayers: an analysis of language models on negation benchmarks}.
\newblock In \emph{Proceedings of the 12th Joint Conference on Lexical and Computational Semantics (*SEM 2023)}, pages 101--114, Toronto, Canada. Association for Computational Linguistics.

\bibitem[{Truong et~al.(2022)Truong, Otmakhova, Baldwin, Cohn, Lau, and Verspoor}]{truong-etal-2022-another}
Thinh~Hung Truong, Yulia Otmakhova, Timothy Baldwin, Trevor Cohn, Jey~Han Lau, and Karin Verspoor. 2022.
\newblock \href {https://aclanthology.org/2022.aacl-main.65} {Not another negation benchmark: The {N}a{N}-{NLI} test suite for sub-clausal negation}.
\newblock In \emph{Proceedings of the 2nd Conference of the Asia-Pacific Chapter of the Association for Computational Linguistics and the 12th International Joint Conference on Natural Language Processing (Volume 1: Long Papers)}, pages 883--894, Online only. Association for Computational Linguistics.

\bibitem[{van Son et~al.(2016)van Son, van Miltenburg, and Morante}]{van-son-etal-2016-building}
Chantal van Son, Emiel van Miltenburg, and Roser Morante. 2016.
\newblock \href {https://aclanthology.org/W16-5007} {Building a dictionary of affixal negations}.
\newblock In \emph{Proceedings of the Workshop on Extra-Propositional Aspects of Meaning in Computational Linguistics ({E}x{P}ro{M})}, pages 49--56, Osaka, Japan. The COLING 2016 Organizing Committee.

\bibitem[{Wiegand et~al.(2010)Wiegand, Balahur, Roth, Klakow, and Montoyo}]{wiegand-etal-2010-survey}
Michael Wiegand, Alexandra Balahur, Benjamin Roth, Dietrich Klakow, and Andr{\'e}s Montoyo. 2010.
\newblock \href {https://aclanthology.org/W10-3111} {A survey on the role of negation in sentiment analysis}.
\newblock In \emph{Proceedings of the Workshop on Negation and Speculation in Natural Language Processing}, pages 60--68, Uppsala, Sweden. University of Antwerp.

\bibitem[{Yang et~al.(2019)Yang, Dai, Yang, Carbonell, Salakhutdinov, and Le}]{yang2019xlnet}
Zhilin Yang, Zihang Dai, Yiming Yang, Jaime~G. Carbonell, Ruslan Salakhutdinov, and Quoc~V. Le. 2019.
\newblock \href {https://proceedings.neurips.cc/paper/2019/hash/dc6a7e655d7e5840e66733e9ee67cc69-Abstract.html} {{XLNet}: Generalized autoregressive pretraining for language understanding}.
\newblock In \emph{Advances in Neural Information Processing Systems 32: Annual Conference on Neural Information Processing Systems 2019, NeurIPS 2019, December 8-14, 2019, Vancouver, BC, Canada}, pages 5754--5764.

\end{thebibliography}
\bibliographystyle{acl_natbib}

\clearpage

\appendix



\section{Model endpoints}
\begin{itemize}
    \item GPT-4: We accessed GPT-4 through the offical API with the name \texttt{gpt-4}. Note that this is different from the GPT-4 turbo model with the name \texttt{gpt-4-1106-preview}.
    \item LLaMA-2-13B: We used the official instruction fine-tuned LLaMA-2-13B available on the HuggingFace hub with the name: \texttt{meta-llama/LLaMA-2-13b-chat-hf}.
    \item Flan-T5-xxl: We used the official xxl version (11.3B) of the Flan-T5 model available on the HuggingFace hub with the name: \texttt{google/flan-t5-xxl}.

\end{itemize}

\section{Details of Affixal Nonce word prediction task}
\label{sec:nonce_prompt}

\paragraph{List of nonce words}

\ex{roagly, vibble, drok, scrop, plard, hif, tepable, plawic, bluth, sprat, flurf}

\paragraph{List of non-negative affixes}
Prefix: \ex{ambi-, aqu-, ast-, aud-, auto-, bi-, bio-, cent-, circum-, co-, cred-, cycl-, dec-, dia-, equ-, geo-, grad-, hydro-, inter-, medi-, mega-, min-, micro-, pan-, semi-, tele-, uni-, tri-}. Suffix: \ex{-able, -al, -ance, -ful, -ian, -ic, -tic, -ile, -ism, -ist, -junct, -ly}

\begin{tcolorbox}
[colback=black!5!white,colframe=black!75!black,title=Nonce]
\scriptsize

\begin{verbatim}

A nonce word is a word occurring, invented, 
or used just for a particular occasion, or a 
word with a special meaning used for a special 
occasion. Infer whether the given nonce word 
contains negation or not.

A word contains negation if it has a negated 
meaning, usually expressed through a negative 
prefix (such as un, in) or suffix (such as 
less). 

The word decentralize contains negation. True 
or False?
Answer: True
Explanation: decentralize is created by 
prepending the root word centralize with the 
negative prefix de.

The word deserve contains negation. True or 
False?
Answer: False
Explanation: deserve just coincidentally starts 
with de.

The word {word} contains negation. True or 
False?
Answer:

\end{verbatim}
\end{tcolorbox}

\section{Prompts for sentiment analysis}
\label{sec:sentiment-prompts}

\begin{tcolorbox}
[colback=black!5!white,colframe=black!75!black,title=Word-level sentiment]
\scriptsize

\begin{verbatim}

{Few-shot samples}

The sentiment of the word {word} is positive, 
negative, or neutral. 

Answer:
\end{verbatim}
\end{tcolorbox}

\begin{tcolorbox}
[colback=black!5!white,colframe=black!75!black,title=Sentence-level sentiment]
\scriptsize

\begin{verbatim}

{Few-shot samples}

{sentence}
Question: Is this sentence positive or 
negative?

Answer: 
\end{verbatim}
\end{tcolorbox}


\section{Full results}


\begin{table}[!htbp]
\footnotesize
    \centering
    \begin{tabular}{P{2.25cm}P{1cm}P{2cm}P{0.5cm}}
    \toprule
    Model  & Neg. Nonce & Non-neg. Nonce & All  \\
    \midrule
       GPT-4 & 0.434 & 1  & 0.717   \\
       LLaMA-2-13B & 0.575  & 0.991   & 0.783\\
       Flan-T5-xxl & 0.627 & 0.964  & 0.795 \\
       \bottomrule
    \end{tabular}
    \caption{Affixal nonce word prediction task}
    \label{tab:nonce}
\end{table}

    

\begin{table*}[!t]
\footnotesize
    \centering
    \begin{tabular}{r ccc | c}
    
     \toprule
       Model & \multicolumn{3}{c}{Accuracy} & \multicolumn{1}{c}{NegMorph} \\
       & Neg & Non-neg & All & Correct  \\
    \midrule
    \multicolumn{1}{l}{\textit{Affix (fine-tuned)}} & \\
    \midrule
        BERT & 0.940 & 0.959 & 0.949 & 0.516 \\
        RoBERTa & 0.931 & 0.964 & 0.947 & 0.614 \\
        AlBERT & 0.933 & 0.956 & 0.945 & 0.737 \\
        XLNet & 0.950 & 0.915 & 0.932 & 0.707 \\
        GPT-2 & 0.928 & 0.949 & 0.938 & 0.614 \\
    \midrule
    \multicolumn{1}{l}{\textit{Affix (zero-shot)}} & \\
    \midrule
       GPT-4 & 0.783 & 0.994 & 0.888  & 0.671    \\
       
       LLaMA-2-13B & 0.707 & 0.770 & 0.738 & 0.658  \\

       Flan-T5-xxl & 0.867  & 0.976 & 0.921 &  0.751  \\
    \midrule
    \multicolumn{1}{l}{\textit{Affix (fewshot})} & \\
    \midrule
       GPT-4 & 0.890 ({\color{teal} \scriptsize +0.107})  & 0.997 ({\color{teal} \scriptsize +0.003}) & 0.943 ({\color{teal} \scriptsize +0.055}) & 0.670  \\
       LLaMA-2-13B & 0.767 ({\color{teal} \scriptsize +0.060})  & 0.938 ({\color{teal} \scriptsize +0.168}) & 0.852 ({\color{teal} \scriptsize +0.114}) &  0.658  \\

       Flan-T5-xxl & 0.855 ({\color{red} \scriptsize -0.012}) & 0.993 ({\color{teal} \scriptsize +0.017})  & 0.924 ({\color{teal} \scriptsize +0.003})  & 0.750  \\
    \midrule
    \multicolumn{1}{l}{\textit{Affix (fewshot)-Hyphen}} & \\
    \midrule
       GPT-4 & 0.916 ({\color{teal} \scriptsize +0.133}) & - & - & 0.929 (\color{teal} \scriptsize +0.258)  \\
       LLaMA-2-13B & 0.956  ({\color{teal} \scriptsize +0.249})  & - & - & 0.984 (\color{teal} \scriptsize +0.326)  \\
       Flan-T5-xxl & 0.948 ({\color{teal} \scriptsize +0.081})  & - & - & 0.968 (\color{teal} \scriptsize +0.217)  \\
    
    \bottomrule
    \end{tabular}
    \caption{Results of our affixal negation prediction task. ({\color{teal}+}/{\color{red}-} denote the change compared to the \textit{Affix (zero-shot)} setting}
    \label{tab:accuracy}
\end{table*}



\end{document}